\documentclass[conference]{IEEEtran}
\IEEEoverridecommandlockouts
\usepackage{cite}
\usepackage{xspace}
\usepackage[linesnumbered]{algorithm2e}
\usepackage{amsmath,amssymb,amsfonts}
\usepackage{algorithmic}
\usepackage{graphicx}
\usepackage{hyperref}
\newcommand{\SYST}{\textsc{CREAT}e\textsc{-IR}\xspace}
\newcommand{\create}{\textsc{CREAT}e\xspace}
\usepackage{textcomp}
\usepackage{xcolor}
\def\BibTeX{{\rm B\kern-.05em{\sc i\kern-.025em b}\kern-.08em
    T\kern-.1667em\lower.7ex\hbox{E}\kern-.125emX}}

\begin{document}

\title{\create: Clinical Report Extraction and Annotation Technology}

\author{
Yichao Zhou$^\dag$, Wei-Ting Chen$^\dag$, Bowen Zhang$^\dag$, David Lee$^\dag$, J. Harry Caufield$^\ddag$, \\ Kai-Wei Chang$^\dag$, Yizhou Sun$^\dag$, Peipei Ping$^\ddag$ and Wei Wang$^\dag$\\
$^\dag$ Department of Computer Science, University of California, Los Angeles. \\
$^\ddag$ Departments of Physiology, Medicine and Bioinformatics, University of California, Los Angeles.\\
\fontsize{9}{9}\selectfont\ttfamily\upshape
\{yz,weitingtw,bzhang0527,dkmlee,kwchang,yzsun,weiwang\}@cs.ucla.edu; \\
\{jcaufield,pping\}@mednet.ucla.edu\\
}

\maketitle

\begin{abstract}
Clinical case reports are written descriptions of the unique aspects of a particular clinical case, playing an essential role in sharing clinical experiences about atypical disease phenotypes and new therapies. 
However, to our knowledge, there has been no attempt to develop an end-to-end system to annotate, index, or otherwise curate these reports.
In this paper, we propose a novel computational resource platform, \create, for extracting, indexing, and querying the contents of clinical case reports. \create fosters an environment of sustainable resource support and discovery, enabling researchers to overcome the challenges of information science. An online video of the demonstration can be viewed at \url{https://youtu.be/Q8owBQYTjDc}.
\end{abstract}

% \begin{IEEEkeywords}
% \end{IEEEkeywords}

\section{Introduction}

% There is a perennial need to automatically and precisely curate the clinical case reports into structured knowledge, i.e. extract important clinical named entities and relationships from the narratives~\cite{aronson2010overview,savova2010mayo,soysal2018clamp,,alfattni2020extraction}.
Case reports are a time-honored means of sharing observations and insights about novel patient cases~\cite{Caban-Martinez2012,caufield2019comprehensive}. 
%caufield2018reference
As of 2020, at least 160 case report journals were in existence, with over 90\% having open access policies and almost half indexed by PubMed~\cite{mcentyre2001pubmed}. 
The narrative of a case report details the symptoms, diagnosis, treatment, and outcome of an individual, describing observations made over the course of clinical care.

\begin{figure}[ht]
\centering
\includegraphics[width=6cm]{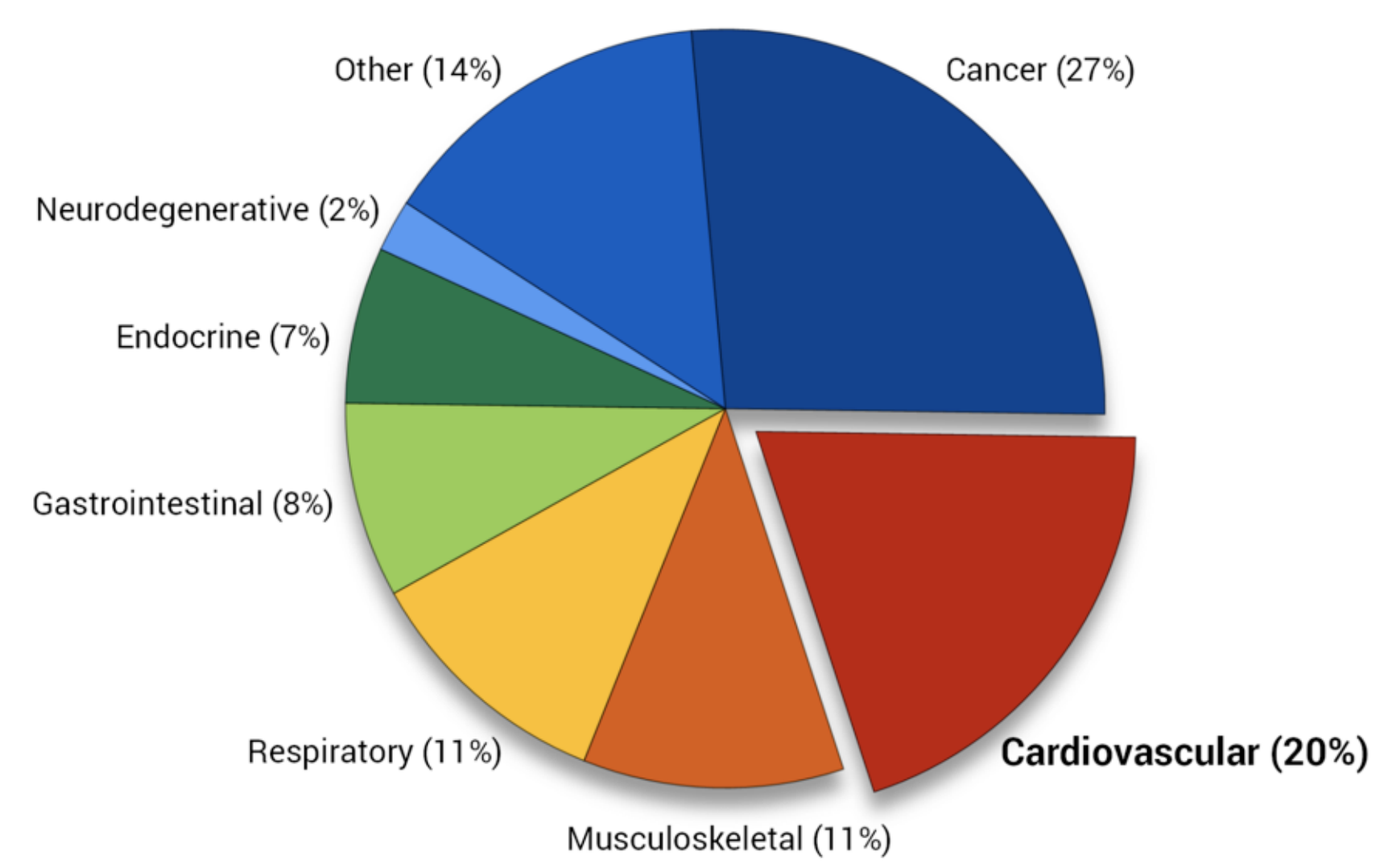}
\caption{Cardiovascular disease accounts for 20\% of all
case reports, and is the 2nd largest category of case reports after cancer.}
\label{fig:cardio}
\end{figure}

Often these case reports contain exceptionally valuable clinical data, addressing unusual disease situations. To our knowledge, there has been no attempt to annotate, index, or otherwise curate these reports. Unlike other types of medical literature, there are no organizational frameworks or methodical review articles for case reports, and no metadata standards exist for their curation. A significant challenge exist for this rapidly growing corpus. Focusing our efforts, we address the domain of cardiovascular disease, an important area for which a range of research and clinical questions occur frequently. As shown in Figure~\ref{fig:cardio}, cardiovascular disease accounts for 20\% of all case reports. %, and is the 2nd largest category of case reports after cancer.

In this paper, we demonstrate an end-to-end systems incorporating algorithms for extracting, indexing, and querying the contents of clinical case reports. Our proposed system, Clinical Report Extraction and Annotation Technology (\create), automates generation of metadata about case reports, unlocking this important data resource through a searchable portal. 

To make case reports findable and accessible, our primary innovation is a graph-based representation of each case report, where nodes signify concepts described in the narrative (e.g. sign/symptoms, diagnosis, etc). To build a case report's graph, we start by employing named entity recognition techniques to identify concepts (nodes), which are then standardized against existing biomedical ontology~\cite{caufield2019comprehensive} to make the metadata interoperable. Concepts are then connected (edges) by detecting described temporal relationships in the text, facilitating retrieval over the patient chronologies that are hallmarks of case report descriptions. Collectively, \create framework will be used to share metadata around published case reports, making it reusable by others via public APIs. 

In this demo, we present four key properties of our system. 
\begin{itemize}
    \item As of October 2020, it offers rich resources of over 10k reports for cardiovascular disease with depositions from a wide range of sources such as Scientific Literature and Authorized User Submissions, as shown in Figure~\ref{fig:architecture}. 
    \item It provides a PDF submission service, based on Grobid~\cite{lopez2009grobid}, which is able to convert the publications in PDF format into well organized XML format. Metadata such as title, author, affiliation information can be automatically extracted for users by text mining technology.
    \item It is powered by \SYST, a relation-based information retrieval system for clinical case reports, which outperforms solr~\cite{smiley2015apache}. Instead of simple keyword match, \SYST embeds advanced deep learning algorithms to extract significant named entities and relations from the narrative. Relative case reports are retrieved from the database based on these structured knowledge. This also enables a temporal reasoning on the user queries and provides better search results.
    \item It provides a user-friendly interface for entity and relation annotation. A graph visualization of temporal order of the clinical events is generated for each document. 

\end{itemize}

\section{System Architecture and Design}
\begin{figure}[htbp]
\centering
\includegraphics[width=4cm]{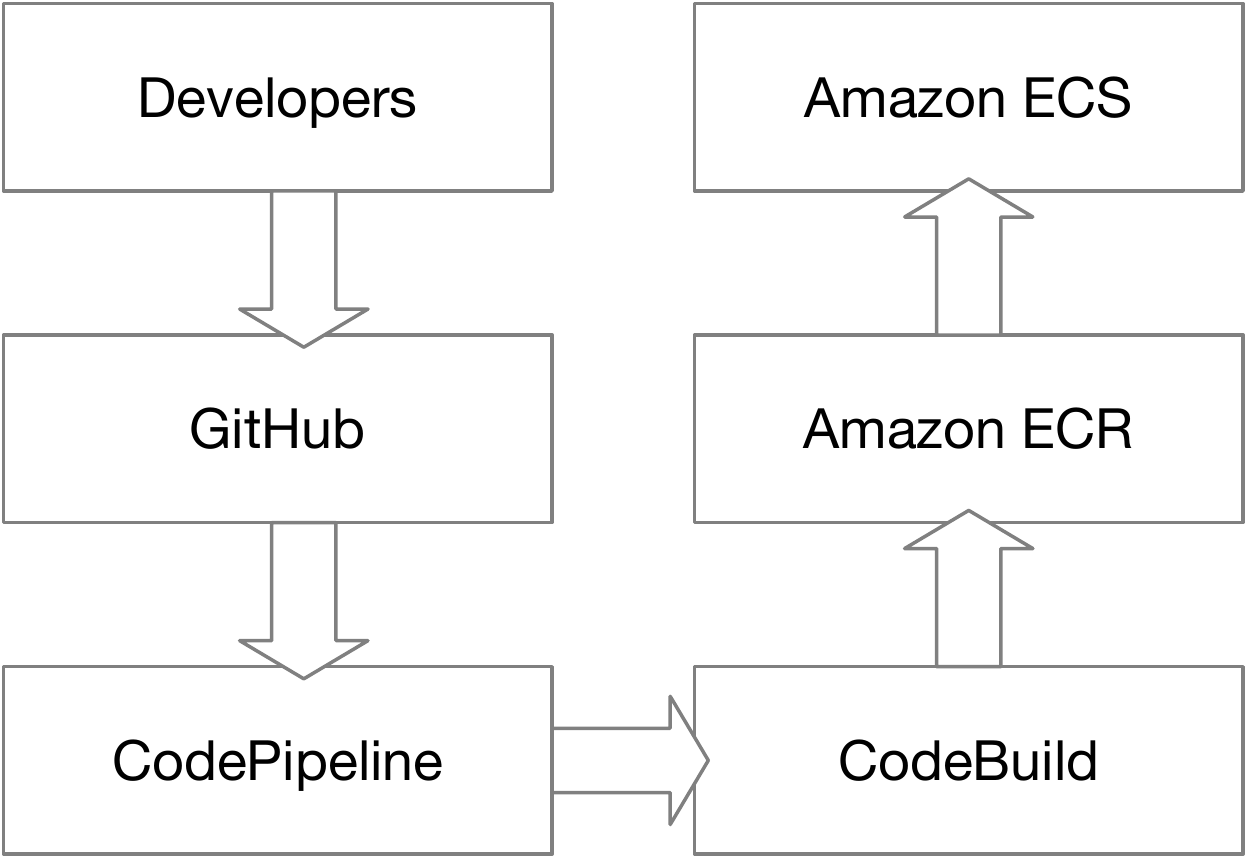}
\caption{The Development Pipeline of \create}
\label{pipeline}
\end{figure}

\create is a cloud-based application and its service is mainly hosted on the Amazon ECS (Fargate) with a CI/CD pipeline as illustrated in Figure \ref{pipeline}.

\begin{figure}[htbp]
\centering
\includegraphics[width=8.5cm]{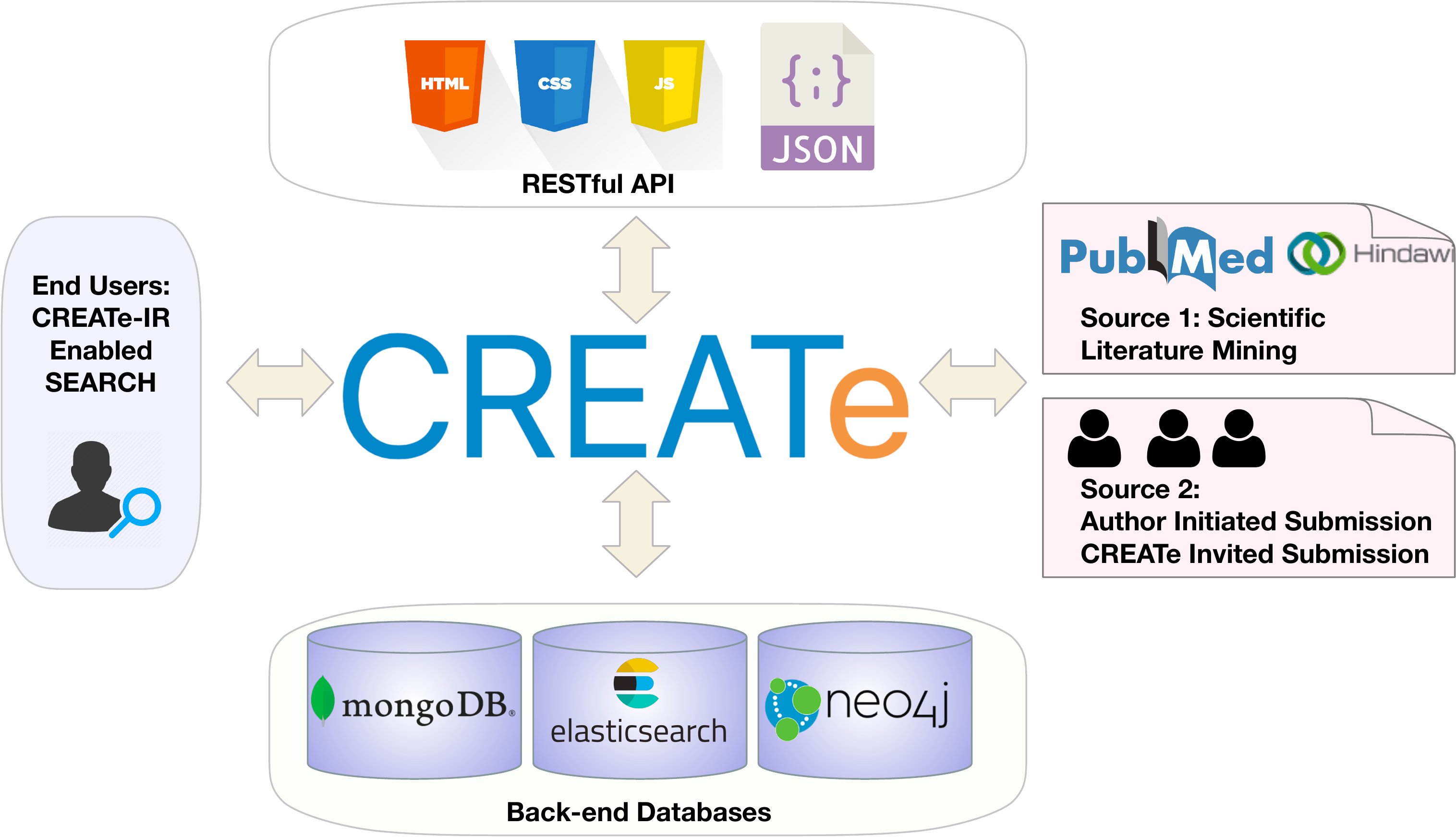}
\caption{The Architecture of CREATe}
\label{fig:architecture}
\end{figure}

As shown in Figure \ref{fig:architecture}, the main feature of \create is to allow users perform \SYST search, which will be elaborated in later sections. The two sources of data are case reports collected from PubMed and user-input reports. The frontend of \create is a single page application developed using React, which allows users to communicate with the RESTful API in the backend built with Express. The application is served by Nginx, a light-weight software for web serving. The majority of data for \create, is stored in the MongoDB server for persistency. The data in MongoDB server is queried via the Express backend to ensure security and consistency. This is the same for both Neo4j server and ElasticSearch server, which allow the application to perform complex search as explained in later section.

\section{\SYST}
\subsection{Data Source and Preprocessing}
Identification of data source begins with a query to PubMed using the publication type and MeSH term filters to locate cardiovascular (CVD) case reports. A query of PubMed in six areas of CVD (cardiomyopathy, ischemic heart disease, cerebrovascular accidents, arrhythmias, congenital heart disease, and valve disease) returns around 118,000 case reports. From these results, a web crawler (e.g. Apache Nutch) is used to locate the associated case reports and publication metadata. The contents can be captured in XML or online PDFs. The XML and PDF documents will be parsed into plain text to facilitate subsequent analysis, organized into case report sections and sentences.

\subsection{Data Annotations}
\begin{figure*}[htbp]
\centering
\includegraphics[width=18cm]{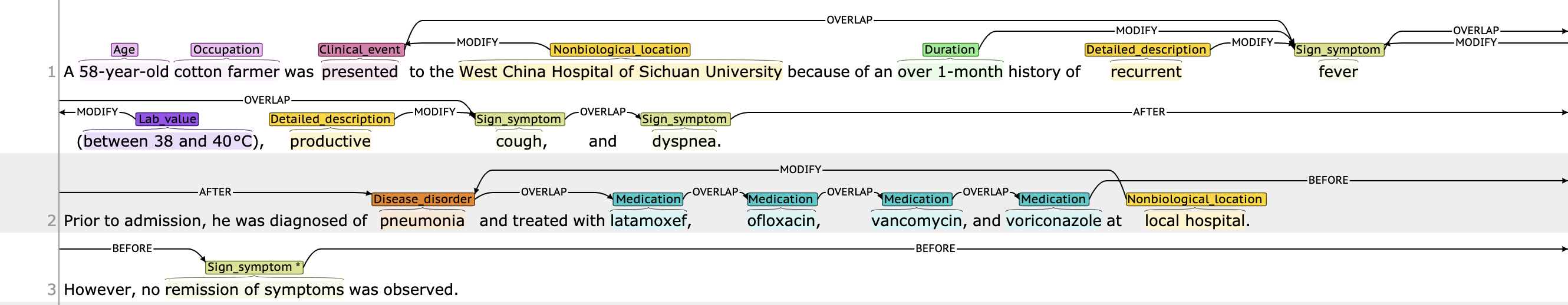}
\caption{Example of the BRAT annotation interface, showing labeled textual elements linked by temporal and semantic relational structures.}
\label{fig:brat}
\end{figure*}

To enable a supervised learning process for the automatic extraction of events/entities and relations from the clinical case reports, we build an embedded interface (Figure~\ref{fig:brat}) for creating, editing, and visualizing document annotations based on the brat rapid annotation tool (BRAT)~\cite{stenetorp2012brat}. BRAT provides basic annotation primitives through the assignment of type labels to spans of text. These base annotations can be connected by directed or undirected edges using standard user interface gestures into \textit{n}-ary associations, enabling the representation of sophisticated relational structures.
In addition, we develop a comprehensive typing schema for information extraction from clinical narratives~\cite{caufield2019comprehensive} and invite several medical experts to annotate hundreds of case reports based on this schema. 
Details are explained as following:
\begin{itemize}
    \item We annotate EVENTS which are text elements representing situations or conditions and trigger a progression in a patient’s clinical course (e.g. \textit{dyspnea} as \textit{Sign/Symptom}). 
    \item We annotate ENTITIES which are general non-trigger text elements which play a semantic role in the clinical narrative (e.g. \textit{cotton farmer} as \textit{Occupation}). 
    \item We also annotate RELATIONS which define the associations between two EVENTS, or between EVENTS and ENTITIES. Relations include temporal relations, such as BEFORE, AFTER, and OVERLAP, which order events in time. Furthermore, they include semantic relations, such as IDENTICAL and MODIFY, which reflect the meaning between words.
\end{itemize}

\subsection{Information Extraction from User Queries}
Once we receive a user query like ``A patient was admitted to the hospital because of fever and cough.'', we apply two machine learning modules to parse the query and extract important knowledge from them (i.e. hospital (\textit{Non-biological Location}), fever (\textit{Sign/Symptom}), cough (\textit{Sign/Symptom}), and the temporal relation between cough and fever (OVERLAP)). 

\noindent \textbf{Named Entity Recognition.}
We first develop a machine learning module, called named entity recognizer, using contextualized token representations to locate and classify clinical terminologies into predefined categories, such as diagnostic procedure, disease disorder, severity, medication, medication dosage, and sign symptom. 
In specific, we pre-train a deep contextualized model C-FLAIR\footnote{The pre-trained model can be download at \url{https://drive.google.com/drive/folders/1b8PQyzTc\_HUa5NRDqI6tQXz1mFXpJbMw?usp=sharing}.} with millions of parameters, in order to provide rich token embeddings for knowledge extraction. The training process is developed on one NVIDIA Tesla V100 GPU (16GB) and takes one week to finish. We conduct experiments with three public datasets to test C-FLAIR and find our approach outperforms the state-of-the-art methods by 1.5\% on average F1 score in terms of the extraction performance. 

\noindent \textbf{Temporal Relation Reasoning.}
We continue to predict the temporal relations among the extracted named entities. There is a consensus with the clinical community regarding the difficulty of temporal information extraction, due to the high demand for domain knowledge and high complexity of clinical language representations. 
\begin{figure}[htbp]
\centering
\includegraphics[width=6.5cm]{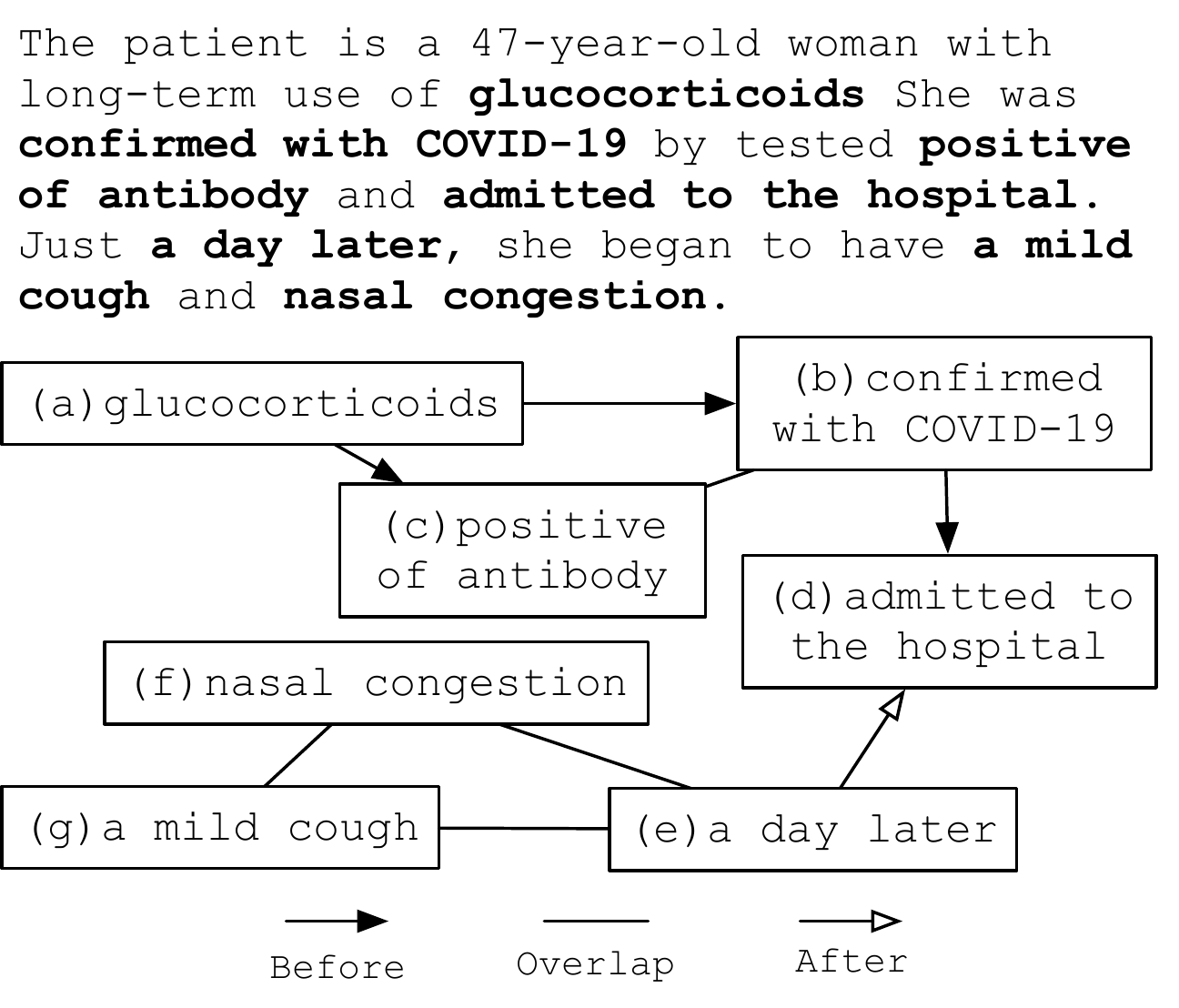}
\caption{An illustration of a clinical case report with  its partial temporal graph where transitivity dependencies exist.}
\label{fig:example}
\end{figure}
We notice the dependencies among events in one clinical document are key enabler of classifying the temporal relations. For example in Figure~\ref{fig:example}, given that \texttt{b} happened before \texttt{d}, \texttt{e} happened after \texttt{d} and \texttt{e} happened simultaneously with \texttt{f} , we can infer according to the temporal transitivity rule that \texttt{b} was before \texttt{f}.
However, existing methods either require expensive feature engineering or are incapable of modeling the global relational dependencies among the events. In our system, we build a temporal relation extraction module~\cite{zhou2020clinical} based on common dependencies such as transitivity and symmetry patterns. % as shown in Table~\ref{tab:psl}. 
Specifically, as we predict a series of temporal relations from one clinical report, we compute a score to measure the satisfaction of all dependencies among these predicted relations, in order to generate an extra loss term to regularize the model training process. Extensive experiments on two public dataset, I2B2-2012~\cite{sun2013evaluating} and TB-Dense~\cite{cassidy-etal-2014-annotation} show that our temporal relation extraction approach significantly outperforms baseline methods by 1.98\% and 2.01\% per F1 score.

\if 0
\begin{table}[t]
    \centering
    \caption{Temporal transitivity and symmetry dependencies. $A,B,C$ are three terms representing the clinical events.}
    \resizebox{.7\linewidth}{!}{
    \begin{tabular}{|l|}
    \hline
        Transitivity Dependencies \\
        \hline
        Before$(A,B)$ $\land$ Before$(B,C)$ $\rightarrow$ Before$(A,C)$ \\
        Before$(A,B)$ $\land$ Overlap$(B,C)$ $\rightarrow$ Before$(A,C)$ \\
        Overlap$(A,B)$ $\land$ Before$(B,C)$ $\rightarrow$ Before$(A,C)$ \\
        Overlap$(A,B)$ $\land$ Overlap$(B,C)$ $\rightarrow$ Overlap$(A,C)$ \\
        After$(A,B)$ $\land$ After$(B,C)$ $\rightarrow$ After$(A,C)$ \\
        After$(A,B)$ $\land$ Overlap$(B,C)$ $\rightarrow$ After$(A,C)$ \\
        Overlap$(A,B)$ $\land$ After$(B,C)$ $\rightarrow$ After$(A,C)$ \\
        \hline
        \hline
        Symmetry Dependencies \\
        \hline
        Before$(A,B)$ $\rightarrow$ After$(B,A)$ \\
        After$(A,B)$ $\rightarrow$ Before$(B,A)$\\
        Overlap$(A,B)$ $\rightarrow$ Overlap$(B,A)$ \\
    \hline
    \end{tabular}}
    \label{tab:psl}
\end{table}
\fi
\begin{figure}[htbp]
\centering
\includegraphics[width=7cm]{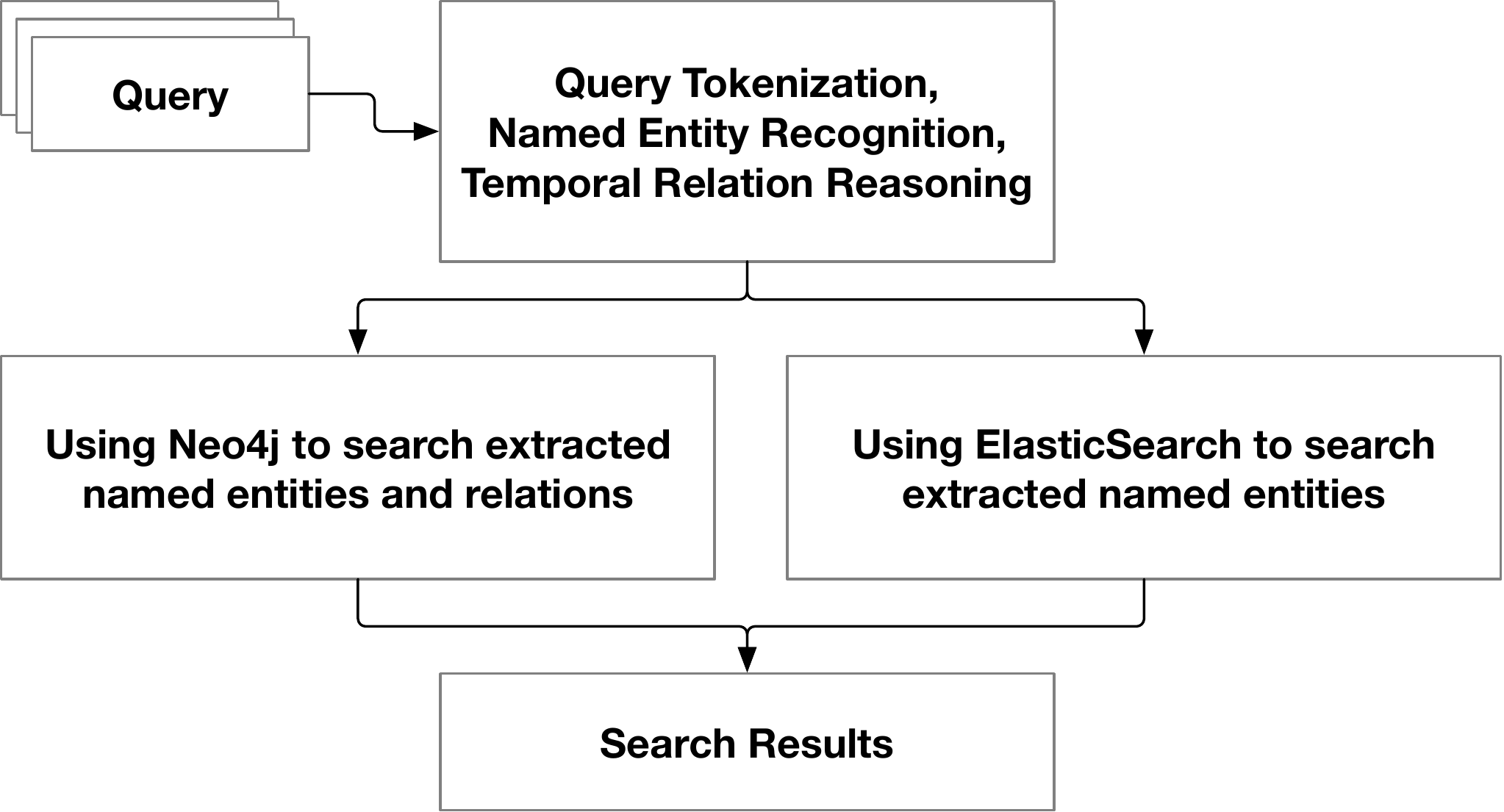}
\caption{The search workflow of \SYST}
\label{fig:search}
\end{figure}
\begin{figure*}[h]
\centering
\includegraphics[width=16cm]{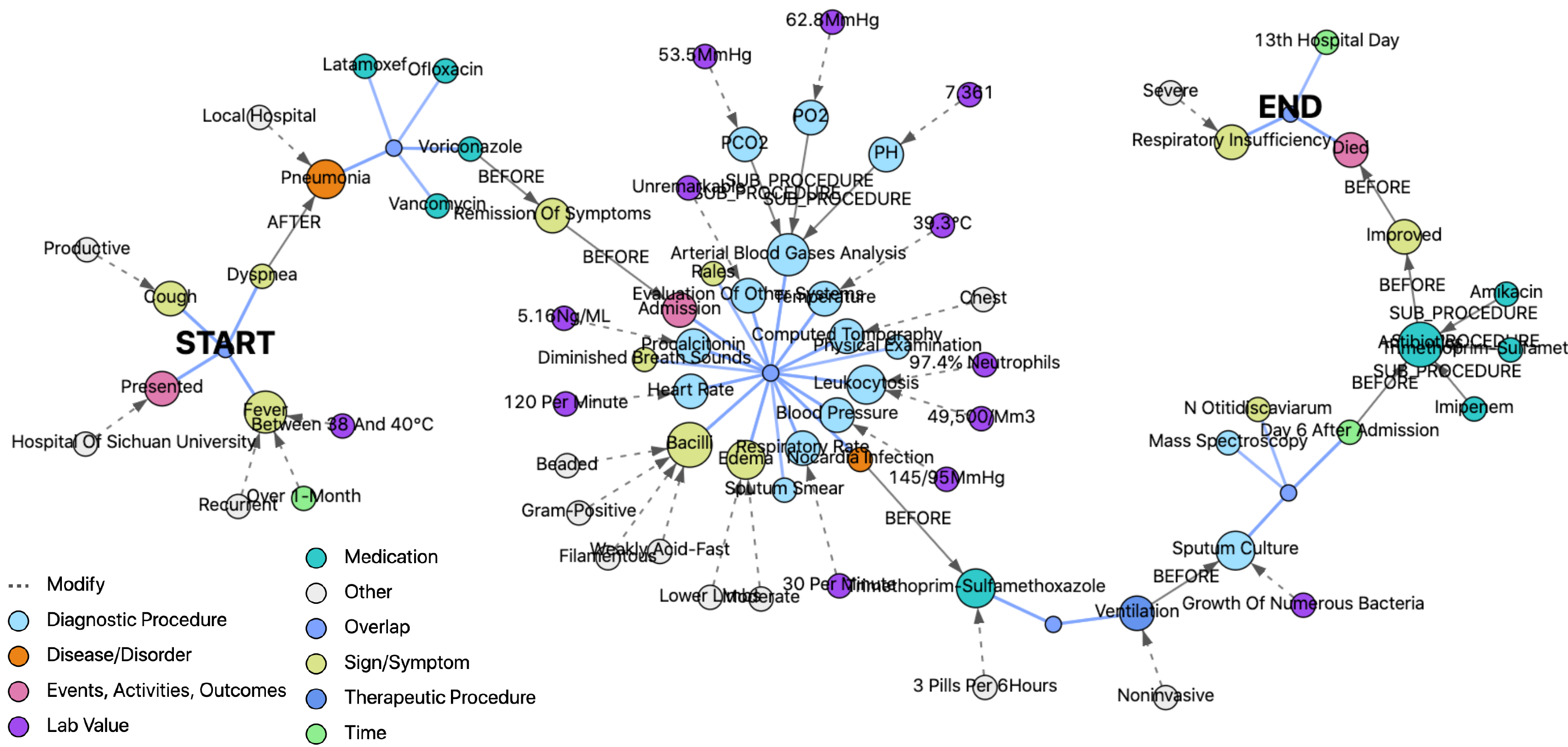}
\caption{Example network graph visualization representing a clinical case matching the query: ``A patient was admitted to the hospital because of fever and cough''. The start of the graph reflects that the patient presented overlapping symptoms of fever and cough, thus aligning with the temporal and semantic structure of the query. The nodes and edges that follow reflect the evolution of the patient’s case, which ultimately ended in death due to respiratory failure.}
\label{network_graph}
\end{figure*}

\subsection{Search Approach}

We provides multiple search functions in our search engine, including search by keywords, search by entities and search by relations. ElasticSearch and Neo4j are utilized to build up the search component, where ElasticSearch mainly handles keyword search and Neo4j handles entity\&relation search. For independent resource occupation purposes, a collection of case reports are indexed separately on each search engine.  
Figure~\ref{fig:search} illustrates the overall search flow. % that occurs from when user inputs his query to when the search result is returned. 
By default, Neo4j is the primary search engine in \SYST system. The results returned by Neo4j will be placed on top, followed by results from ElasticSearch. 

\noindent \textbf{Neo4j. }
In Neo4j, data is saved as a graph of nodes and edges. Therefore, indexing each case report into Neo4j requires a transformation from texts to nodes and edges. A particular node will contain a \textit{nodeId}, a \textit{label} and a \textit{entityType}. Property \textit{label} keeps a natural language description for the node and property \textit{entityType} represents the classification of this node. %Following shows an example of a node in the database. 
%$$\{nodeID:'T6', label: 'pain', entityType: 'Sign\_symptom'\} $$
An edge will contain a \textit{source}, a \textit{target} and a \textit{label}, which records the source nodeId and destination nodeId, as well as the relation type. Then, all nodes and edges are put into Neo4j via cypher query.

\noindent \textbf{ElasticSearch. }
To better cater to the keyword search demand in \SYST system, we build the document index with customized analyzer. In ElasticSearch, an analyzer can be divided into three sub-components: token filters, tokenizers and character filters. For token filter, we choose \textit{asciifolding}, \textit{lowercase}, \textit{snowball}, \textit{stop} and \textit{stemmer}. For tokenizer, considering that some of the symptoms or medications may have longer names, we select N-gram tokenizer and customize it with min\_gram=3 and max\_gram=25.
%With analyzer properly configured and all the documents are indexed in, \SYST is ready to handle keyword search requests via its ElasticSearch component.

\subsection{Event Visualization in Temporal Order}
Annotations break the text structure down to event anchors and the temporal relations between them. These representations emphasize the atomic units and relational structures that are crucial for an understanding of the progression of a clinical case. \SYST represents these underlying structures through network graph visualizations, which show the interconnections between sets of entities and events, giving focus to the semantic roles played by these fine-grained elements over the course of a clinical narrative. An example of such a visualization is shown in Figure \ref{network_graph}. This visualization component is rendered using scalable vector graphics under a force-directed algorithm, which distributes nodes and clusters in space to minimize their repulsive energies and crossing edges. This component also possesses functionality for recognizing standard user interface gestures. The layout of nodes can be reconfigured by selecting a node with the mouse and dragging it to a different location in space. Similarly, the visualization window can be adjusted by zooming and panning using mouse wheel and drag gestures, respectively.

\section{Conclusion}
In this paper, we demonstrate \create, the first end-to-end system for annotating, indexing and curating the clinical case reports. \create is powered by \SYST, which includes two state-of-the-art information extraction techniques to parse the important clinical events and temporal relationships for retrieving related clinical documents. We also embed a comprehensive annotation system and PDF parsing module to our system in order to facilitate further data collection. %Our future work includes making \create interoperable with other data resource indexing platform and expansible to other scientific domains beyond cardiovascular disease.

\section{Acknowledgement}
We would like to thank the anonymous reviewers for their helpful comments. We also want to thank Shunji Zhan, Zixia Weng and Don Lee for the discussion. This work was supported by NIH R35 HL135772.

\bibliographystyle{IEEEtran}
\bibliography{reference} 

\end{document}